\documentclass[preprint,12pt]{elsarticle}

\usepackage{amssymb}
\usepackage{amsmath}
\usepackage{subfig}
\usepackage{float}

\journal{Physica D: Nonlinear Phenomena}

\begin{document}

\begin{frontmatter}



\title{A Distribution-to-Distribution Neural Framework for Probabilistic Forecasting of Nonlinear Dynamical Systems}


\author[1]{Tianlin Yang}

\author[1,2,4]{Hailiang Du\corref{cor1}}
\ead{hailiang.du@durham.ac.uk}

\author[1]{Louis Aslett}

\affiliation[1]{
    organization={Department of Mathematical Sciences, Durham University},
    city={Durham},
    postcode={DH1 3LE},
    country={United Kingdom}
}

\affiliation[2]{
    organization={Institute of Hazard, Risk and Resilience, Durham University},
    city={Durham},
    postcode={DH1 3LE},
    country={United Kingdom}
}

\affiliation[4]{
    organization={Data Science Institute and Global School of Sustainability, London School of Economics and Political Science},
    city={London},
    postcode={WC2A 2AE},
    country={United Kingdom}
}

\cortext[cor1]{Corresponding author}

\begin{abstract}
Probabilistic forecasts provide a principled representation of forecast uncertainty through predictive distributions. Directly propagating predictive distributions with conventional physical models requires solving distributional evolution equations, a task well-known to be computationally prohibitive for high-dimensional nonlinear systems. Consequently, multi-step predictive distributions are commonly approximated using ensembles or sampled trajectories. A novel distribution-to-distribution (D2D) neural probabilistic forecasting framework is proposed with compatible distributional encoding and decoding interfaces around a replaceable neural forecasting module. Kernel mean embeddings represent input distributions, while mixture density networks parameterise outputs, enabling recursive multi-step forecasting in distributional form. The framework is demonstrated on the Lorenz63 system. Results show that the D2D model captures nontrivial distributional evolution, produces skillful probabilistic forecasts without explicit ensemble simulation, and remains competitive with a naively initialised perfect-model benchmark. This provides proof-of-concept evidence that predictive distributions can be recursively propagated through a neural forecasting model and evaluated in terms of probabilistic forecast performance.
\end{abstract}

\begin{graphicalabstract}
\includegraphics[width=\textwidth]{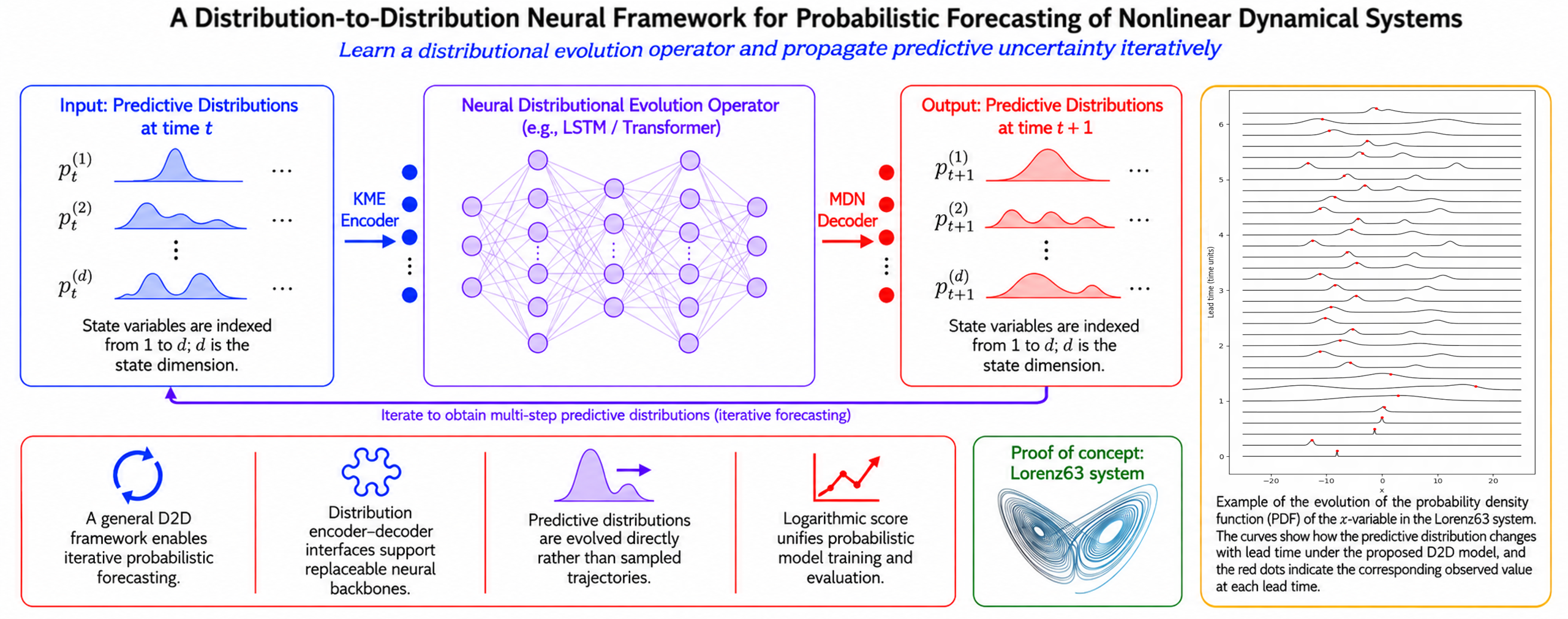}
\end{graphicalabstract}

\begin{highlights}
\item A general D2D framework enables iterative probabilistic forecasting.
\item Predictive distributions are recursively propagated without trajectory sampling.
\item Logarithmic score unifies probabilistic model training and evaluation.
\item Distribution encoder-decoder interfaces support replaceable neural backbones.
\item Skillful forecasts are obtained without explicit ensemble simulation.
\end{highlights}


\end{frontmatter}



\section{Introduction}

Prediction of complex systems across science and engineering is often formulated in terms of dynamical systems governed by ordinary or partial differential equations, grounded in physical knowledge of the underlying processes. Such physics-based models underpin a wide range of applications, including geophysical flows, climate and weather prediction, engineering control systems, and biological dynamics. Alongside these approaches, data-driven methods have long been explored as alternative or complementary tools for forecasting dynamical systems. Recent advances in machine learning have led to significant progress in data-driven forecasting, particularly neural network based forecasters, which achieve predictive skill comparable to, and in some cases exceeding, that of physics-based models, while offering substantial gains in computational efficiency \citep{bi2023accurate, lam2023learning, kochkov2024neural, marino2016building}. Despite their different formulations, both physics-based and existing neural network based forecasters share a fundamental structural limitation: they evolve single system states forward in time, producing deterministic trajectories rather than explicitly evolving predictive distributions. 

As is well known, propagating full distributions under dynamical models generally requires solving associated density-evolution equations, such as the Liouville equation for deterministic dynamics or the Fokker--Planck equation for stochastic dynamics, a task that is computationally intractable for high-dimensional nonlinear systems \citep{Ruelle1989, MajdaHarlim2012}. This structural limitation poses a fundamental challenge for uncertainty quantification. In complex, nonlinear and often chaotic systems, uncertainty arises from multiple sources, including for example observational noise, unresolved processes, and model discrepancy. Quantifying and propagating this uncertainty is essential for both scientific understanding and decision-making, particularly when forecasts are used to assess risk, threshold exceedance, or low-probability high-impact events.

A wide range of uncertainty quantification approaches have been developed, with ensemble-based methods forming the backbone of practical uncertainty representation in many forecasting systems. Initial condition ensembles are commonly used to represent uncertainty arising from imperfect observations and data assimilation. Parameter ensembles, often referred to as perturbed-physics ensembles, aim to capture uncertainty associated with model parameterisation. Multi-model ensembles further attempt to address structural model discrepancy by combining forecasts from models with different formulations. In practice, these ensemble strategies are designed to address different sources of uncertainty in a rather independent manner. When combined, their joint use can incur substantial computational cost and may lead to overly dispersed predictive distributions, thereby reducing the practical value of the forecast.

Beyond ensemble-based approaches \citep{leith1974theoretical, toth1997ensemble}, a variety of other uncertainty quantification methods have been developed, including Bayesian inference and filtering techniques \citep{evensen2003ensemble, kennedy2001bayesian}, stochastic parameterisations \citep{buizza1999stochastic}, and post hoc statistical calibration methods \citep{gneiting2007probabilistic}. While these approaches provide valuable tools for characterising uncertainty in specific settings, they generally do not formulate forecast evolution directly as a mapping between predictive distributions.

Similar uncertainty quantification strategies have also been widely adopted in machine-learning-based forecasting, particularly for neural network models.
Data-driven probabilistic forecasting and uncertainty quantification for nonlinear dynamical systems have also been explored using Gaussian-process, generative, and ensemble-based approaches \cite{gu2024probabilistic,maulik2023quantifying}.
Ensemble-based approaches are commonly used to represent predictive uncertainty by training multiple networks with different initialisations, data subsets, or perturbations, mirroring ensemble techniques in physics-based modelling. In addition, a range of methods have been developed specifically for neural networks, including approximate Bayesian techniques such as dropout-based inference \citep{gal2016dropout}, as well as more recent generative approaches based on diffusion and probabilistic latent variable models \citep{kingma2013auto,price2025probabilistic}. While these models can produce rich probabilistic forecasts for a single forward step, multi-step forecasting typically relies on propagating sampled trajectories rather than explicitly evolving the predictive distribution itself.


Probabilistic forecasting provides a principled and natural framework for
uncertainty quantification by representing predictions as full probability
distributions, which offer the most complete description of predictive uncertainty. The importance of probabilistic forecasting has been increasingly
recognised across the broader forecasting literature, both in methodological
developments and in large-scale empirical forecasting studies
\citep{petropoulos2022forecasting, makridakis2022m5}. Crucially, the quality
of probabilistic forecasts can be evaluated using proper scoring rules, such as
the logarithmic score \citep{good1952rational}. In this setting, probabilistic
forecast skill provides a principled criterion for evaluating the quality of the
associated uncertainty quantification.

From this perspective, uncertainty quantification is no longer an auxiliary diagnostic, but should be treated as a primary modelling objective that is explicitly optimised during training. Achieving this objective in dynamical forecasting requires predictive uncertainty to be propagated coherently through time, so that the evolution of the system is described in terms of predictive distributions rather than individual trajectories. However, existing dynamical forecasting models, whether physics-based or neural-network-based, remain fundamentally trajectory-based.

In this work, a distribution-to-distribution (D2D) neural probabilistic forecasting framework is proposed, allowing predictive distributions themselves to be treated as primary objects that are propagated through the forecasting model. More specifically, the methodological question addressed in this work is whether a neural probabilistic forecaster can be constructed such that its output predictive distribution can be used directly as the input to the same forecasting model at the next step, thereby enabling recursive multi-step forecasting in distribution space without reverting to trajectory sampling. The proposed framework addresses this problem by introducing a neural architecture with compatible distributional input and output representations, thereby creating a closed forecasting loop in which predictive distributions can be propagated recursively from one forecast step to the next. Input distributions are encoded using kernel mean embeddings and processed by a neural forecasting module, while the outputs are decoded into parameterised predictive distributions via mixture density networks. By making predictive uncertainty an intrinsic component of the model evolution, the framework also allows model training and optimisation to be driven explicitly by probabilistic performance metrics.

The remainder of the paper is organised as follows. Section 2 describes the proposed D2D framework, including the output distribution parameterisation, the kernel-based encoding of input distributions, the logarithmic-score training objective, and the associated training strategies. Section 3 presents a proof-of-concept experimental demonstration based on the Lorenz63 system. Section 4 concludes the paper with a summary of the main results and a discussion of limitations and future extensions.

\section{D2D Framework}

The proposed distribution-to-distribution (D2D) framework treats predictive distributions, rather than point states, as the fundamental objects of dynamical forecasting. In this work, predictive distributions are represented through marginal distributions for individual state variables, although the same framework can in principle be extended to joint predictive distributions over the full state vector. The defining feature of the D2D architecture is that both its inputs and outputs are represented in distributional form. Each predictive distribution is represented by a collection of marginal distributions over the $d$ input variables. Crucially, the predictive distribution produced at one forecast step is represented in a form compatible with the model input, allowing it to be recursively fed back for subsequent forecast steps.

\begin{figure}[H]
    \centering
    \includegraphics[width=1\linewidth]{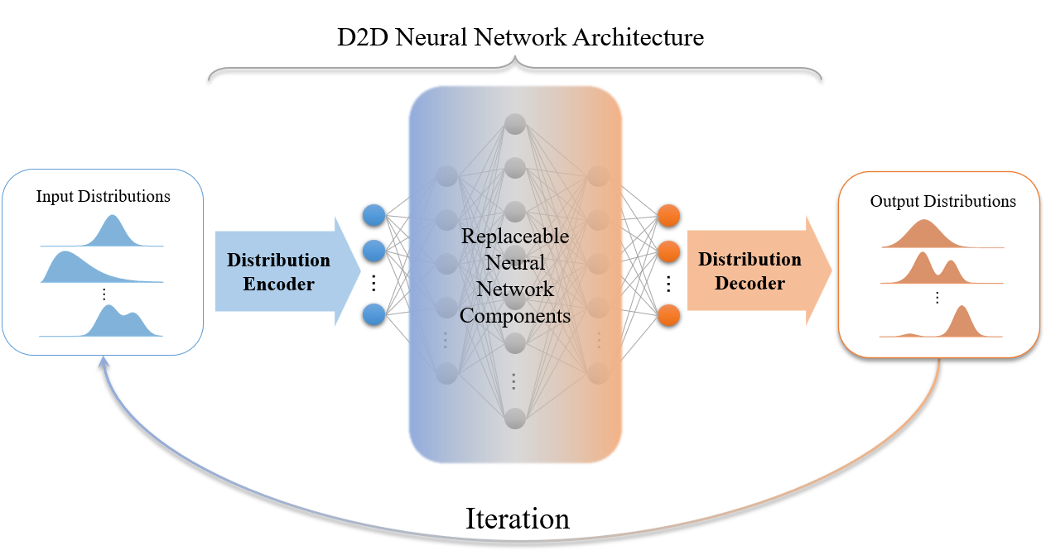}
    \caption{Schematic illustration of the proposed D2D neural network architecture.}
    \label{fig:D2D}
\end{figure}

Fig.~\ref{fig:D2D} illustrates the overall D2D neural network architecture. The framework consists of three main components:
\begin{enumerate}
    \item A \textbf{distribution encoder}, which maps input distributions into a finite-dimensional representation suitable for neural processing;
    \item A \textbf{replaceable neural forecasting module}, which transforms the encoded representation through standard neural architectures (e.g., LSTM \citep{hochreiter1997long},  Transformer \citep{vaswani2017attention}, or other backbone models);
    \item A \textbf{distribution decoder}, which maps the neural outputs back to parameterised predictive distributions.
\end{enumerate}

This modular design allows existing neural forecasting architectures to be incorporated into the D2D pipeline with minimal structural modification. The key innovation lies not in replacing neural backbone models, but in redefining their input and output interfaces so that predictive distributions become primary objects. The present implementation focuses on marginal predictive distributions for each state dimension, mainly for computational efficiency and clarity. This is not a structural restriction of the proposed D2D framework: the same distributional encoding and decoding principle can, in principle, be extended to joint predictive distributions over the full state vector. The main additional cost is the growth of distributional parameters needed to represent dependence structures, with full covariance representations scaling quadratically with the state dimension. A systematic study of joint D2D forecasting is left for future work.


The following subsections detail the four central technical components of the framework: (i) the parameterisation of neural network outputs as predictive distributions, (ii) the encoding of input distributions for neural processing, (iii) the choice of loss function used to train the network, and (iv) the training strategy used for distributional forecasting.

\subsection{From Neural Network Outputs to Predictive Distributions}
\label{sec:d2d_output_distribution}

One of the key features of the proposed distribution-to-distribution (D2D) framework is its ability to produce predictive distributions directly. Existing probabilistic forecasting approaches often rely on ensembles (finite samples) to represent predictive uncertainty, with continuous predictive densities obtained only through post-processing when needed. While widely used, ensemble-based approaches exhibit several limitations as a general framework for uncertainty quantification in dynamical forecasting: predictive uncertainty is represented only through a finite ensemble, rather than as an explicit predictive distribution; multiple forward evaluations of the forecasting model are required, increasing computational cost; and uncertainty representation remains an auxiliary layer added on top of the forecasting model rather than an intrinsic part of it, so that its quality is not directly optimised during training. 

More recent neural probabilistic forecasting methods include diffusion-based generative models \citep{price2025probabilistic, li2023seeds} and discretisation-based approaches \citep{espeholt2022deep,andrychowicz2023deep}. Diffusion models can capture complex and multimodal uncertainty, but typically rely on iterative sampling procedures and remain computationally demanding for dynamical forecasting. Discretisation-based methods are straightforward to implement and train, but they yield piecewise-constant predictive distributions whose resolution is constrained by the bin design. Achieving finer granularity requires a larger number of bins, which increases model complexity and computational cost, and can also lead to data sparsity in individual bins.

These considerations motivate the use of mixture density networks (MDNs) \citep{bishop1994mixture} as an integral component of the proposed architecture, enabling predictive distributions to be produced directly by the model outputs. Rather than generating a single point prediction, the network outputs the parameters of a mixture distribution in a single forward pass, providing a flexible and computationally efficient framework for probabilistic forecasting. Formally, given a multivariate input $\mathbf{x}$, the MDN outputs a set of parameters
\[
\{\pi_i(\mathbf{x}), \mu_i(\mathbf{x}), \sigma_i(\mathbf{x})\}_{i=1}^M,
\]
corresponding to the mixture weights, means, and standard deviations of an $M$-component Gaussian mixture. In the present formulation, this parameterisation is used for the predictive marginal distribution of a single target state variable. The predictive marginal density is then given by
\begin{equation}
p(y \mid \mathbf{x}) = \sum_{i=1}^M \pi_i(\mathbf{x}) \,
\mathcal{N}\!\left(y \mid \mu_i(\mathbf{x}), \sigma_i^2(\mathbf{x})\right),
\label{eq:marginal-mdn}
\end{equation}
where the mixture weights $\pi_i(\mathbf{x})$ are constrained to be positive and sum to one via a softmax activation, and the standard deviations $\sigma_i(\mathbf{x})$ are enforced to be positive using an exponential activation. Although Eq.~\eqref{eq:marginal-mdn} is written for a marginal predictive density, the same MDN construction can be extended directly to joint predictive densities by using multivariate mixture components. For a \(d\)-dimensional state, full covariance parameterisation introduces \(d(d+1)/2\) covariance parameters per component, so the decoder output dimension scales quadratically with \(d\). Diagonal or low-rank covariance structures can reduce this cost, at the price of restricting the dependence structure. The use of Gaussian mixtures is also theoretically motivated by their approximation flexibility, since finite Gaussian mixtures can approximate arbitrary target densities arbitrarily closely in $L^1$ as the number and parameters of the components vary \citep{NguyenEtAl2023}.


Conceptually, MDNs are closely related to kernel dressing \citep{broecker2008ensemble}, as both represent predictive distributions as finite mixtures of kernel functions, typically Gaussians. The crucial distinction lies in parameterisation: kernel dressing uses fixed kernel parameters applied uniformly across samples, whereas MDNs learn the mixture weights, means, and variances as nonlinear functions of the input through the neural network itself. In the present setting, this flexibility enables MDNs to capture non-Gaussian and multimodal predictive marginal distributions with minimal additional computational overhead relative to modern neural network architectures. 


\subsection{Encoding Input Distributions for Neural Processing}

A key technical challenge in distribution-to-distribution learning is how to represent input distributions in a form suitable for neural network processing. While mixture density networks enable flexible parameterisation of output distributions, the model must also accept distributions as inputs. More importantly, the distributions produced by the MDN decoder must themselves be re-encoded and fed back into the network, so that predictive distributions can be iteratively propagated across forecast steps. This requires a principled and computationally tractable representation of predictive distributions in a finite-dimensional space.

In this work, kernel mean embedding (KME) \citep{smola2007hilbert, sriperumbudur2010hilbert} is adopted to encode input distributions. Given a positive definite kernel $k:\mathcal{X}\times\mathcal{X}\to\mathbb{R}$ defined on the state space $\mathcal{X}$, the kernel mean embedding of a probability distribution $P$ over $\mathcal{X}$ is defined as
\[
\mu_k(P) := \mathbb{E}_{X \sim P}[\,k(X, \cdot)\,],
\]
where \(X\) is a random variable with distribution \(P\). The resulting object \(\mu_k(P)\) lies in the reproducing kernel Hilbert space (RKHS) associated with the kernel \(k\) \citep{aronszajn1950theory}. Under characteristic kernels (e.g., the Gaussian RBF kernel used in this work), this embedding uniquely represents the distribution \citep{sriperumbudur2010hilbert,muandet2017kernel}. In practice, the embedding is evaluated at a finite set of centres $\{c_j\}_{j=1}^{n_c}$, yielding a finite-dimensional representation
\[
\mathbf{z}_P = \big( \mu_k(P)(c_1), \dots, \mu_k(P)(c_{n_c}) \big)^\top,
\]
which serves as the input to the neural forecasting module. In the present work, a one-dimensional Gaussian radial basis function (RBF) kernel is employed for the predictive marginal distribution of each state dimension:
\[
k(x,c) = \exp\!\left(-\frac{(x-c)^2}{2\ell^2}\right),
\]
where $\ell>0$ is the kernel bandwidth parameter. When the input distribution is Gaussian, \(P=\mathcal{N}(\mu,\sigma^2)\), the embedding admits a closed-form expression, obtained by evaluating a standard Gaussian integral (equivalently, the convolution of two Gaussians) \citep{smola2007hilbert, muandet2017kernel}:
\[
\mu_k(P)(c)
= \sqrt{\frac{\ell^2}{\ell^2+\sigma^2}}
\exp\!\left(
-\frac{(\mu-c)^2}{2(\ell^2+\sigma^2)}
\right).
\]
This closed-form expression enables efficient and exact evaluation of the embedding at the selected kernel centres for Gaussian predictive distributions, without resorting to numerical integration or Monte Carlo approximation.

The kernel bandwidth $\ell$ plays a critical role in controlling the smoothness and sensitivity of the embedding. 
In classical kernel-based methods, kernel hyperparameters such as the Gaussian bandwidth are often chosen using heuristic rules, for example in maximum mean discrepancy (MMD)–based applications \citep{gretton2012kernel}. While such choices are often effective in generic distribution-comparison settings, they are not tailored to the predictive forecasting task considered here. In the present framework, the bandwidth is treated as a learnable parameter. For each state dimension, $\ell$ is optimised jointly with the neural network parameters. This adaptive treatment allows the embedding representation to align with the scale and variability of predictive uncertainty learned from data.

A further crucial property of kernel mean embeddings is their linearity with respect to the underlying distribution:
\[
\mu_k\!\left(\sum_i w_i P_i\right)
= \sum_i w_i \mu_k(P_i).
\]
This linearity plays a central role in the proposed framework. Since the MDN decoder produces predictive distributions in the form of Gaussian mixtures, the embedding of a mixture distribution can be computed analytically as the weighted sum of the embeddings of its individual Gaussian components. Consequently, the encoding is naturally matched to the MDN-based output representation, ensuring that input and output distributions are expressed within the same functional form. This alignment enables seamless iteration of the D2D mapping, allowing predictive marginal distributions to be propagated consistently through successive forecasting steps. It is worth noting that neither the mixture distribution produced by the MDN nor the kernel choice is restricted to the Gaussian form. While Gaussian mixtures combined with Gaussian RBF kernels provide analytical convenience and computational efficiency, the D2D framework only requires structural compatibility between the chosen mixture distribution family and the kernel function. 

The same compatibility also holds in the multivariate case. For a multivariate Gaussian distribution and a multivariate Gaussian RBF kernel, the kernel mean embedding admits a closed-form expression, and by linearity the embedding of a multivariate Gaussian mixture is again given by the weighted sum of the embeddings of its Gaussian components. Therefore, extending the D2D framework to joint predictive distributions does not require Monte Carlo re-encoding in principle. The main practical issue is the selection and number of kernel centres in the joint state space. A tensor-product grid of centres would grow exponentially with the state dimension \(d\), whereas learned, random, or adaptively selected centres provide more scalable alternatives. More general multivariate extensions and theoretical properties of kernel embeddings are discussed in \citep{muandet2017kernel, briol2025dictionary}.


\subsection{The Importance of Using Logarithmic Score}

Recent advances in machine learning have led to remarkable and highly visible success in classification tasks, including spam detection, image recognition, and medical diagnosis. Many of the most influential modern AI systems are built upon classification-based formulations at their core. For example, large language models perform next-token prediction as a multi-class classification problem, and game-playing systems such as AlphaGo rely on classification-based architectures in key decision-making components \citep{radford2019language, brown2020language, silver2017mastering}. In contrast, the impact of machine learning on regression and continuous-valued prediction problems has been comparatively less prominent.

One important distinction between classification and regression lies in the way predictive models are typically formulated and trained. In classification, models output categorical probability distributions and are commonly trained using cross-entropy, which, for standard one-hot targets, is equivalent to the negative log-likelihood. By comparison, regression is often formulated in terms of deterministic point prediction, with mean squared error (MSE) commonly used as the training objective. Note that minimising MSE is equivalent to minimising the negative log-likelihood of Gaussian predictive distributions with constant variance. In practice, nonlinear dynamics are the norm rather than the exception in real-world systems, and uncertainty propagated through such nonlinear mechanisms rarely preserves a Gaussian structure. As a result, the true conditional predictive distribution is seldom well approximated by a Gaussian form.

It is important to note that there exist many other possible ways to evaluate probabilistic predictions. For example, accuracy, the Brier score \citep{glenn1950verification}, and AUC are commonly used in classification \citep{hanley1982meaning, bradley1997use}, while the continuous ranked probability score (CRPS) is widely adopted in continuous-valued forecasting \citep{matheson1976scoring, gneiting2007strictly}. In practice, however, cross-entropy has become the de facto standard as a training loss function in classification, not primarily because of its probabilistic interpretation as log-likelihood, but rather because of its computational convenience: its differentiability and seamless integration with gradient-based optimisation have made it the natural choice for training modern classification architectures. At the evaluation stage, alternative performance metrics such as accuracy, Brier score, and AUC remain widely used.

For standard categorical classification, cross-entropy coincides with the negative log-likelihood and, equivalently, with the logarithmic score \citep{good1952rational}, also known as Ignorance in the forecasting literature \citep{roulston2002evaluating}. From a theoretical standpoint, the logarithmic score occupies a uniquely distinguished position: it is the only proper local scoring rule \citep{brocker2007scoring}, meaning that it uniquely rewards “truthful” probability assignments based solely on the realised outcome. As discussed in \citep{du2021beyond}, the logarithmic score possesses several distinctive advantages over alternative proper scoring rules: Unlike the logarithmic score, nonlocal scoring rules such as CRPS can admit “unfortunate” evaluations that may favour forecasts placing negligible or even zero probability mass on the realised outcome; the logarithmic score admits a direct interpretation in terms of probabilities and bits of information, quantifying forecast skill as information gain; and it is the only proper scoring rule invariant under smooth transformations of the forecast variable, ensuring coherent evaluation across equivalent representations of predictive uncertainty. Despite these theoretical properties, the logarithmic score is not uniformly adopted as the primary evaluation metric in practical forecasting applications, where alternative performance measures are often preferred. Somewhat paradoxically, the widespread success of modern AI in classification may owe much to the fact that its training objective coincides with the uniquely principled logarithmic score.

A unified probabilistic perspective implies that the logarithmic score should serve as the standard criterion for both classification and regression tasks, and for both model training and forecast evaluation. While classification already aligns with this principle through the widespread use of cross-entropy as the training loss, regression or continuous-valued forecasting do not consistently adopt the same standard. This motivates the use of the logarithmic score as a common objective for both model training and forecast evaluation in the proposed framework.

Within the proposed D2D framework, this perspective has direct practical implications. Since both inputs and outputs of the model are probability distributions, the natural training objective is the negative log-likelihood of the realised outcome under the predicted distribution. For a collection of \(N\) forecasts with predictive densities \(\{p_i(x)\}_{i=1}^N\) and corresponding observed outcomes \(\{x_i^{\mathrm{obs}}\}_{i=1}^N\), the empirical loss function is defined as
\[
\mathcal{L}
=
-\frac{1}{N}\sum_{i=1}^N \log p_i\!\left(x_i^{\mathrm{obs}}\right),
\]
which corresponds exactly to minimising the empirical logarithmic score. More broadly, this choice sets the standard for evaluating the quality of uncertainty quantification in forecasting.


\subsection{Training the D2D Model}

Within the D2D framework, two training strategies can be adopted for probabilistic forecasting of dynamical systems.

\paragraph{Direct strategy}
A D2D model can be trained directly for a specific lead time $\tau$, learning a mapping from the initial state distribution to the predictive distribution at lead time $\tau$. The initial state distribution represents uncertainty in the initial conditions. The training objective is defined as the empirical logarithmic score over $N$ independently initialised forecasts. Specifically, let $p_{\tau}^{(i)}(x)$ denote the predictive marginal density for the $i$-th forecast at lead time $\tau$, and let $x_{\tau}^{(i),\mathrm{obs}}$ denote the corresponding observed outcome. The loss function is
\[
\mathcal{L}_{\text{direct}} 
= -\frac{1}{N} \sum_{i=1}^{N} 
\log p_{\tau}^{(i)}\!\left(x_{\tau}^{(i),\mathrm{obs}}\right),
\]
which corresponds to minimising the empirical logarithmic score at the target lead time.

\paragraph{Iterative strategy}
A D2D model can also be trained for a smaller time increment $\Delta t$ and then recursively applied to propagate the predictive distribution to longer lead times. In this iterative setting, the training objective accounts for multiple forecast steps. Suppose the model is applied recursively for $K$ steps, corresponding to lead times $k\Delta t$, $k=1,\dots,K$. Let $p_{k\Delta t}^{(i)}(x)$ denote the predictive marginal density for the $i$-th forecast at lead time $k\Delta t$, and let $x_{k\Delta t}^{(i),\mathrm{obs}}$ denote the corresponding observed outcome. The loss is then defined as
\[
\mathcal{L}_{\mathrm{iter}}
=
-\frac{1}{NK}
\sum_{i=1}^{N}
\sum_{k=1}^{K}
\log
p_{k\Delta t}^{(i)}
\left(
x_{k\Delta t}^{(i),\mathrm{obs}}
\right).
\]
which jointly optimises the empirical logarithmic score over multiple lead times.

In practice, training the iterative model can be stabilised using a curriculum learning strategy over lead times \citep{bengio2009curriculum}. The model is first trained for one-step forecasting, and the maximum training lead time is then progressively extended from $\Delta t$ to $2\Delta t, 3\Delta t, \ldots$ by increasing $K$ in stages. At each stage, the same multi-lead time logarithmic score objective is used, allowing the model to gradually learn stable recursive distributional propagation while mitigating optimisation instability caused by early error accumulation. More generally, the contributions from different forecast steps could be weighted differently in the iterative objective. For simplicity, however, the present study adopts uniform weighting across all lead times.



\paragraph{Comparison of the two strategies}

While the iterative approach generally requires greater computational resources during training than the direct strategy, 
it offers several conceptual and practical advantages. Firstly, a single iteratively trained D2D model can generate forecasts for arbitrary lead times \(k\Delta t\) without retraining, whereas the direct strategy requires a separately trained model for each lead time of interest. Secondly, the iterative objective incorporates information from multiple forecast lead times simultaneously, thereby exposing the model to a richer set of dynamical constraints during training. This encourages the learned transformation to approximate the underlying dynamics rather than overfitting to a specific forecast lead time. In contrast, a direct model optimised solely for a fixed lead time $\tau$ learns a lead-time-specific mapping. If such a model is subsequently iterated to produce multi-step forecasts, its performance is expected to degrade substantially, since it was not trained to maintain dynamical consistency across successive applications. From a dynamical systems perspective, the iterative approach is therefore better aligned with the objective of learning a distributional evolution operator, rather than a collection of lead-time-specific predictors.



\section{Experimental Demonstration}

To demonstrate the capability of the proposed D2D framework, and in particular its ability to propagate predictive uncertainty iteratively through distribution-to-distribution mappings, experiments are conducted on the classical Lorenz63 chaotic dynamical system \citep{lorenz1963deterministic}. The Lorenz63 system provides a well-established benchmark for nonlinear and chaotic dynamics, where uncertainty propagation plays a central role due to strong sensitivity to initial conditions. This sensitivity leads to rapid error growth and makes informative long-range forecasting intrinsically challenging, a difficulty that is further exacerbated in real-world applications by unavoidable model discrepancy. Although low-dimensional, the system captures fundamental mechanisms of nonlinear instability and uncertainty growth that are characteristic of many real-world dynamical systems. Consequently, demonstrating distribution-to-distribution learning in this setting provides meaningful evidence of the framework’s potential applicability to broader classes of nonlinear dynamical forecasting problems.

The Lorenz63 system \citep{lorenz1963deterministic} is governed by
\begin{equation}
\begin{aligned}
    \frac{dx}{dt} &= \sigma (y-x), \\
    \frac{dy}{dt} &= x(\rho-z)-y, \\
    \frac{dz}{dt} &= xy-\beta z.
\end{aligned}
\end{equation}
with the standard parameter values
$\sigma=10$, $\rho=28$, and $\beta=8/3$.
The system is simulated numerically using a fourth-order Runge--Kutta scheme with a time step of $0.01$ time units, and the resulting discrete-time numerical system is treated as the underlying true dynamical system in the experiments. Observations are obtained by sampling the simulated trajectories every 0.05 time units and adding independent and identically distributed Gaussian observational noise to each of the three state variables. The resulting noisy observations are then used to construct a probabilistic time series forecasting problem, for which predictions are produced using the proposed D2D neural network.

The internal forecasting module of the D2D framework adopts a standard LSTM architecture \citep{hochreiter1997long}. While more advanced architectures such as bidirectional LSTMs \citep{schuster1997bidirectional}, Gated Recurrent Unit (GRU) \citep{cho2014learning}, or Transformers \citep{vaswani2017attention} could be employed, the purpose of this study is to validate the D2D framework itself rather than to optimise the specific temporal neural architecture. A conventional LSTM is therefore sufficient to demonstrate the feasibility and effectiveness of distribution-to-distribution learning.

The temporal input to the LSTM consists of five consecutive KME-encoded representations. At each forecast step, the MDN decoder outputs a predictive distribution in the form of a Gaussian mixture consisting of $M=5$ components for each of the three state variables. For the kernel mean embedding of each state dimension, the centres $\{c_j\}_{j=1}^{n_c}$ are selected as $n_c=50$ uniformly spaced points between the minimum and maximum values of that variable in the dataset. The number of mixture components $M$ and the number of centres $n_c$ are tunable hyperparameters: increasing $M$ enhances the expressive capacity of the predictive distribution and allows more complex multimodal structures to be represented, whereas increasing $n_c$ yields a finer finite-dimensional representation of the input distribution, at the cost of increased computational expense and model complexity. In the present experiments, $M=5$ and $n_c=50$ were found to provide a suitable balance between flexibility and stability.

At the initial forecast time, the input to the D2D model is specified as a Gaussian distribution for each state variable. The mean is taken as the observed value, and the variance is set according to the observational noise level. Here, the noise level is defined as the ratio between the observational noise standard deviation and the empirical standard deviation of the corresponding state variable. For compatibility with the MDN output representation, this initial Gaussian distribution is represented internally as a mixture of $M$ identical Gaussian components. 

\begin{figure}[htbp]
\centering

\subfloat{%
\includegraphics[width=0.5\linewidth]{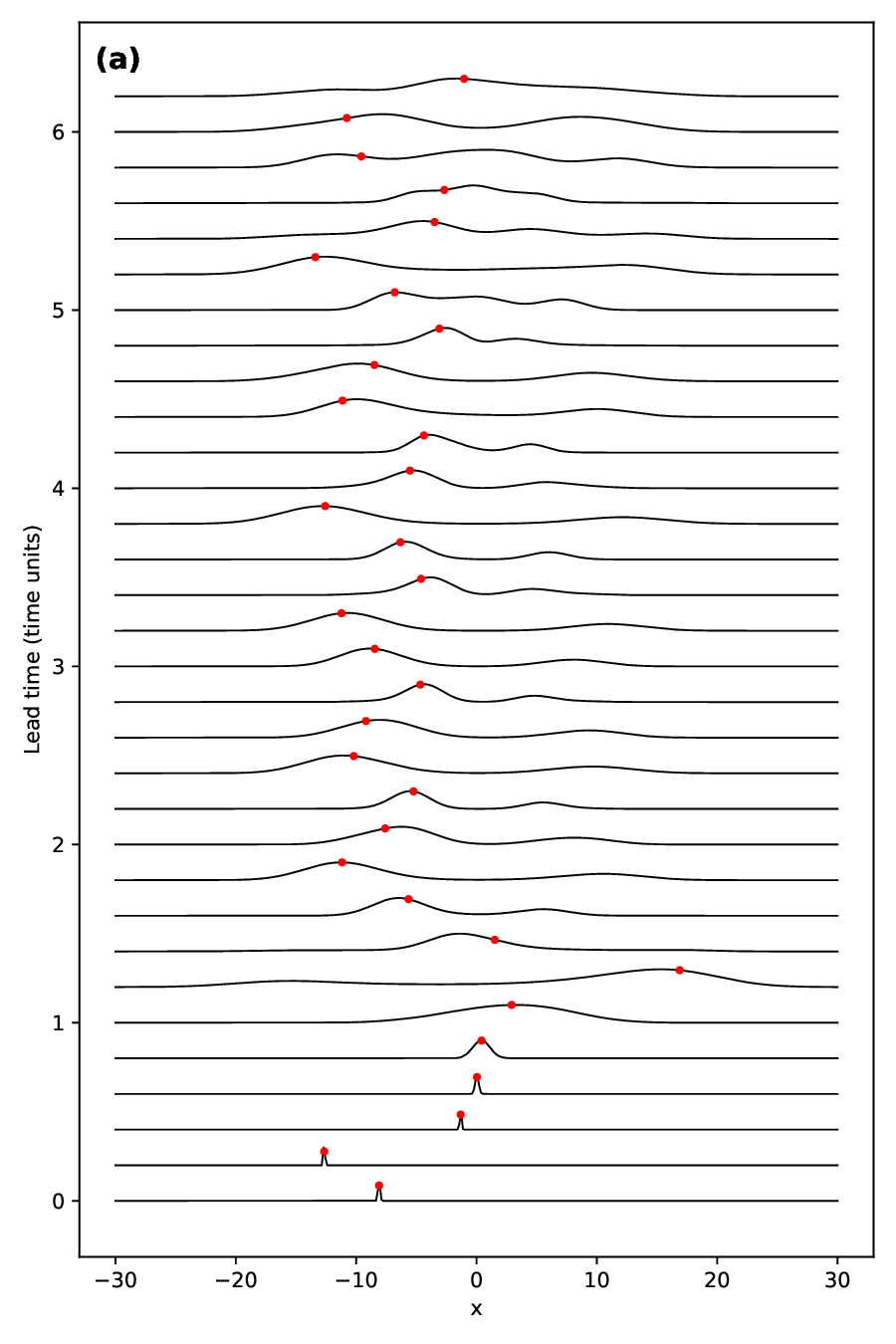}}
\hfill
\subfloat{%
\includegraphics[width=0.5\linewidth]{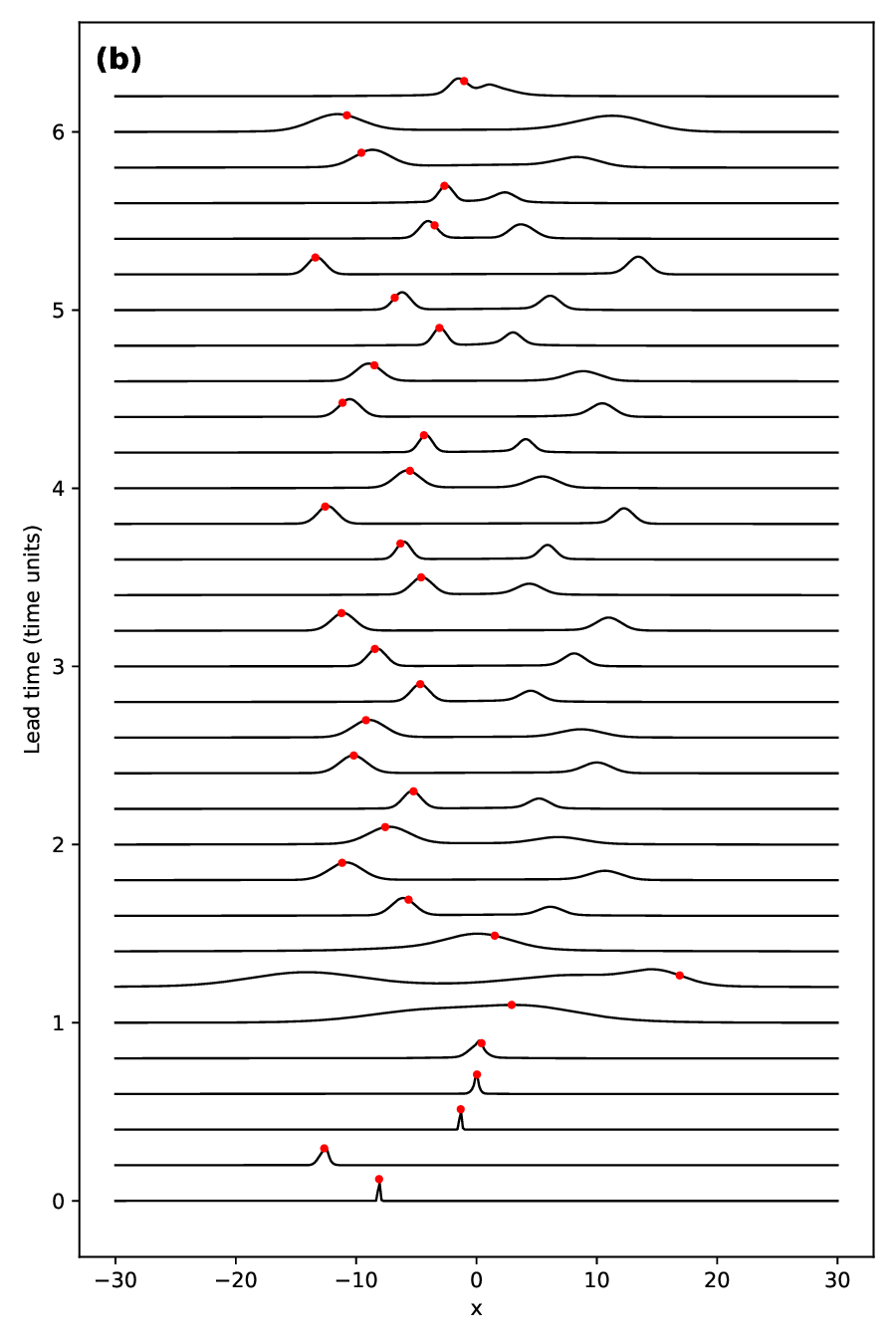}}

\caption{An example of the evolution of the probability density function (PDF) of the $x$-variable. Panel (a) shows the reference distributional evolution under the underlying true dynamical system, approximated empirically by propagating a large ensemble of initial conditions under the numerically simulated Lorenz63 system. Panel (b) shows the corresponding PDF evolution generated by the proposed D2D model trained with the iterative strategy. Red dots indicate the corresponding observed values at each lead time.}
\label{fig:evolve}
\end{figure}

As a first qualitative demonstration of the proposed framework, Fig.~\ref{fig:evolve} visualises an example of the evolution of the initial distribution for the $x$-variable under the underlying true dynamical system in panel (a), and compares it with the corresponding evolution generated by the D2D model in panel (b). The initial distribution at time $0$ is taken to be Gaussian, with mean equal to the observed value and variance determined by the observational noise level, which is set to $0.01$ in this example. The PDF of this initial distribution is propagated forward in time, as shown in Fig.~\ref{fig:evolve}(a), by ensemble simulation using an ensemble of 10{,}000 initial conditions evolved under the numerically simulated Lorenz63 system. The resulting PDF evolution exhibits progressive deformation, stretching, and non-Gaussian structure induced by nonlinear dynamics. Fig.~\ref{fig:evolve}(b) presents the corresponding PDF evolution generated by the proposed D2D model. The model is trained using the iterative strategy with time increment $\Delta t = 0.05$, curriculum expansion up to a maximum lead time of $128 \Delta t =6.4$ time units, and 30{,}000 training samples. The D2D model successfully captures the nonlinear deformation of the predictive distribution over time through recursive distributional propagation. This result demonstrates that the D2D framework is capable of learning a distribution-to-distribution evolution operator, enabling coherent propagation of predictive uncertainty without explicit ensemble simulation.

\begin{figure}[htbp]
\centering

\subfloat{%
\includegraphics[width=0.5\linewidth]{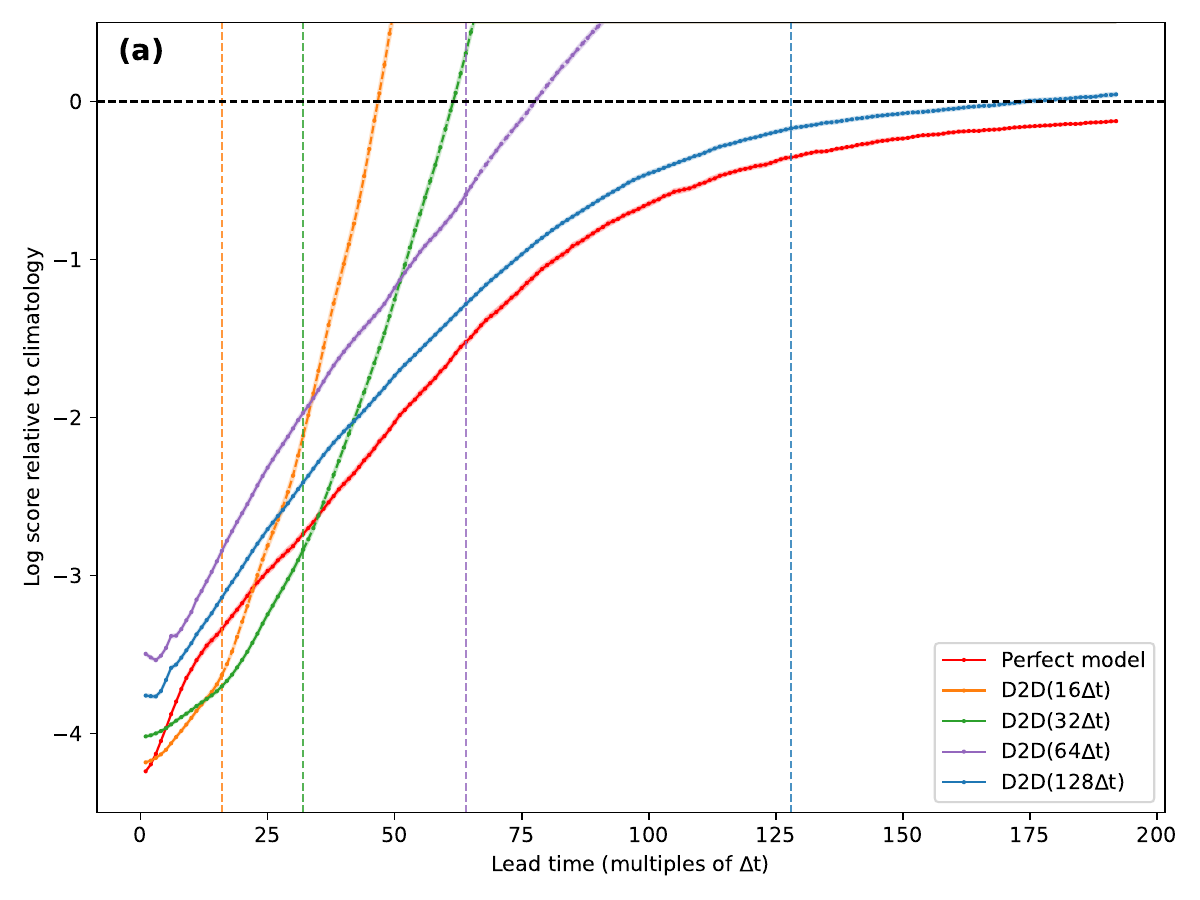}}
\hfill
\subfloat{%
\includegraphics[width=0.5\linewidth]{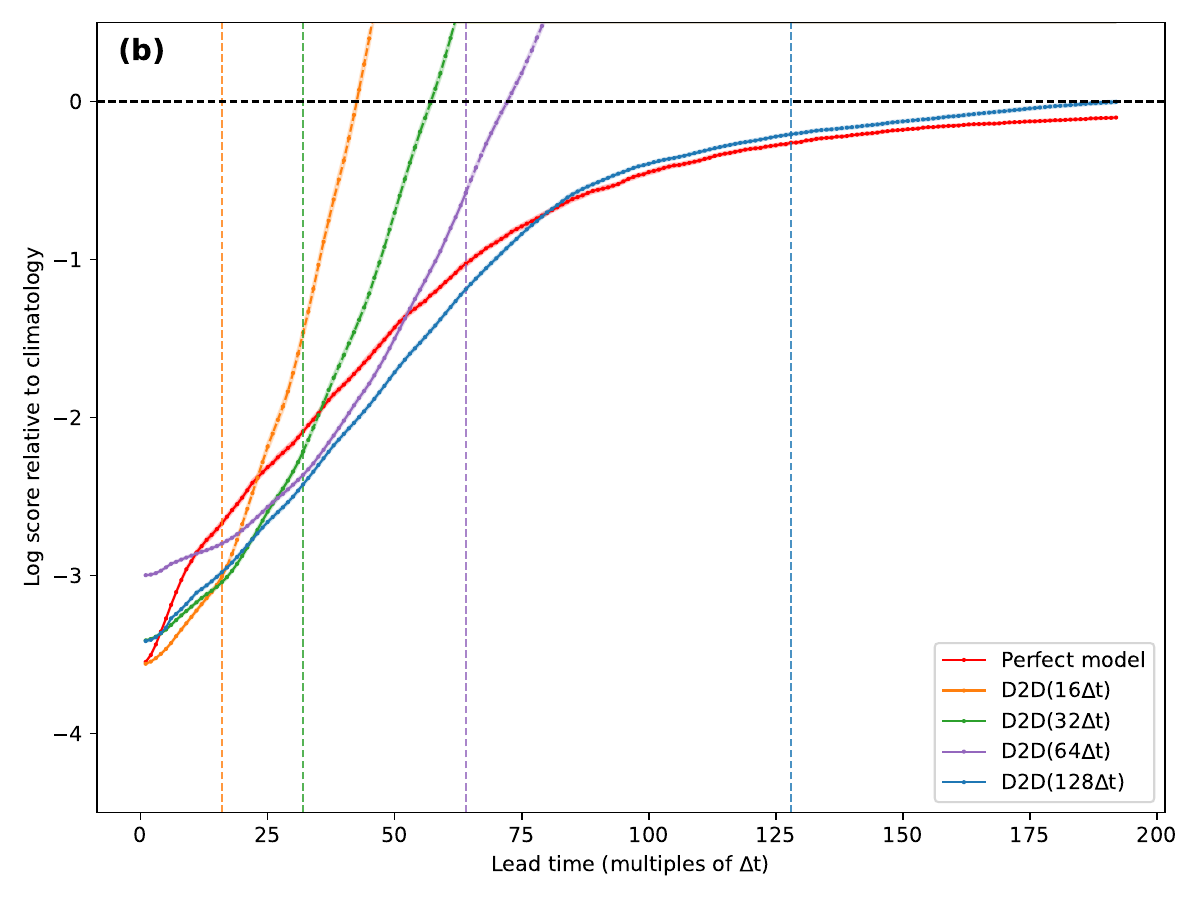}}
\hfill
\subfloat{%
\includegraphics[width=0.5\linewidth]{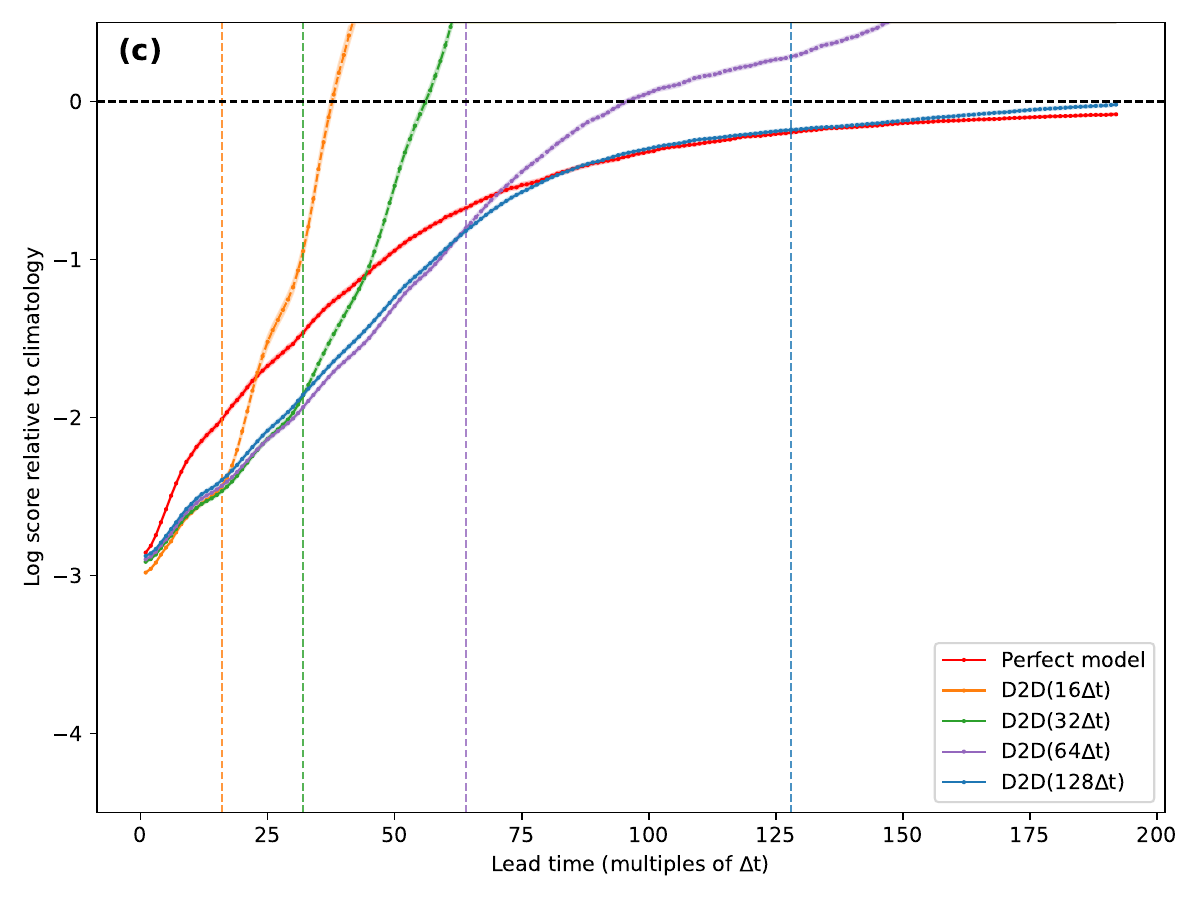}}
\hfill
\subfloat{%
\includegraphics[width=0.5\linewidth]{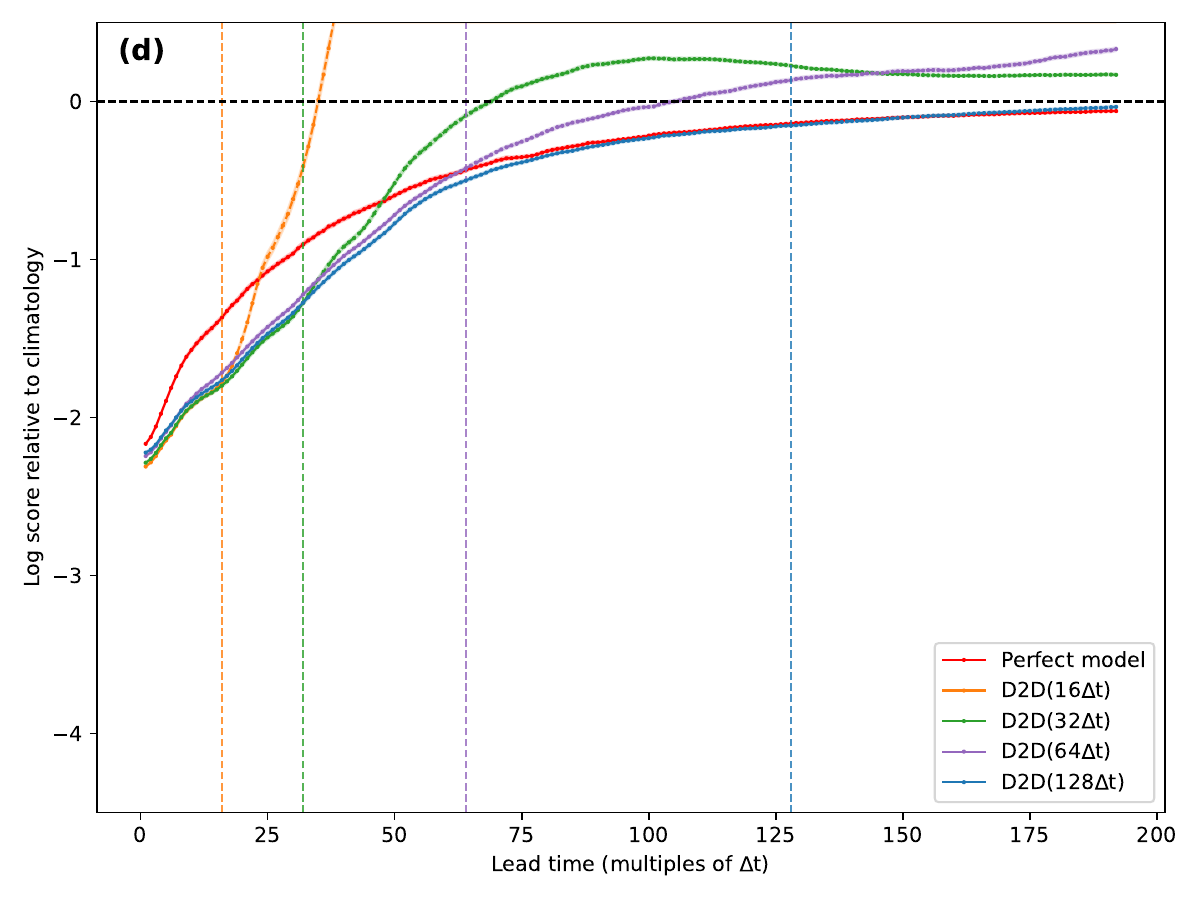}}

\caption{Logarithmic score skill relative to climatology as a function of lead time for iterative D2D forecasts and perfect model forecasts. Panels (a)-(d) correspond to observational noise levels of 0.01, 0.02, 0.04, and 0.08, respectively. The horizontal black dashed line represents the zero skill climatological forecast. The red curve denotes the perfect model forecast based on a 128-member ensemble propagated under the numerically simulated Lorenz63 system and converted to a predictive distribution using Gaussian kernel dressing \citep{roulston2003combining}. The remaining curves correspond to iterative D2D models trained with maximum curriculum lead times of $16\Delta t$, $32\Delta t$, $64\Delta t$, and $128\Delta t$. Shaded bands indicate 95\% bootstrap resampling intervals computed on the test set.}
\label{fig:loss_it_fig}

\end{figure}

To provide a quantitative assessment of predictive skill, iterative D2D forecasts trained with different maximum curriculum lead times are compared against perfect model forecasts constructed from the underlying true dynamical system, as illustrated in Fig.~\ref{fig:loss_it_fig}, where each panel corresponds to a different observational noise level. All forecast performance is evaluated using the logarithmic score relative to the climatological forecast, which serves as the zero-skill reference shown by the horizontal black line in Fig.~\ref{fig:loss_it_fig}. The climatological forecast distribution is obtained by applying kernel dressing to the full set of historical observations, using Gaussian kernels centred at each historical observation and a kernel standard deviation tuned using a large validation set.

For the perfect model forecasts (red curves), an ensemble of 128 initial conditions is propagated forward under the numerically simulated Lorenz63 system, and the resulting ensemble forecast is converted into a predictive distribution at each lead time using kernel dressing. Each kernel is taken to be Gaussian, centred at an ensemble member, with kernel standard deviation tuned separately for each lead time using a large validation set. The remaining curves correspond to iterative D2D models trained with different maximum curriculum lead times, corresponding to $16\Delta t$, $32\Delta t$, $64\Delta t$, and $128\Delta t$. The vertical dashed lines indicate these corresponding maximum curriculum lead times. All models are trained on 30{,}000 observations, with hyperparameters tuned on a separate validation set of 30{,}000 observations, and final evaluation carried out on an independent test set of 30{,}000 observations. To quantify sampling uncertainty in the reported test performance, bootstrap resampling is applied to the test set, and 95\% intervals are shown for each forecast curve. These intervals are barely visible at the plotting scale, suggesting that sampling variability in the reported test performance is very small.

Models trained with short maximum curriculum lead times, such as $16\Delta t$ and $32\Delta t$, show relatively strong performance at short lead times but deteriorate rapidly once forecasts extend beyond the lead-time range covered during training. This is consistent with the fact that these models are explicitly optimised for short term forecast performance and, because they are trained with too few recursive steps, do not access enough dynamical information to learn the underlying evolution. Furthermore, these models do not always achieve the best performance even at their nominal target training lead times. This is particularly evident at low observational noise levels: for example, the $16\Delta t$ model is not consistently optimal at lead time $16\Delta t$, and the $128\Delta t$ model even outperforms the $64\Delta t$ model across all lead times. At larger noise levels, the performance of the $128\Delta t$ and $64\Delta t$ models is also comparable to that of the $16\Delta t$ and $32\Delta t$ models at short lead times. This suggests that capturing the underlying dynamical evolution can become more important than optimising for a shorter target lead time, so longer curriculum training can improve forecast skill even at shorter lead times.

Across the different noise levels shown in Fig.~\ref{fig:loss_it_fig}, the perfect model forecast does not uniformly outperform the D2D models. At low observational noise, it remains a strong benchmark, but as the observational noise level increases, the $64\Delta t$ and $128\Delta t$ models match or surpass the perfect model forecast over almost the entire lead-time range. This is because the perfect model forecast in these experiments is initialised from a simplified probability distribution rather than through a more advanced data assimilation procedure \footnote{In the ideal case, the relevant object would be the perfect initial predictive distribution, although this is generally not available analytically in practice. A corresponding perfect ensemble can in principle be constructed by sampling states that are consistent both with the observational uncertainty and with the long-term dynamics of the system; with a finite observation window, this is more precisely a dynamically consistent ensemble \citep{smith1996accountability}. However, constructing such ensembles is computationally prohibitive and lies beyond the scope of the present study. It should be noted, however, that the D2D model is intended to approximate not only the underlying system evolution but also the predictive effect of such a dynamically consistent initial distribution.}. As a result, the influence of observational noise on the perfect model forecast becomes progressively stronger as the noise level increases. From this perspective, the fact that the D2D models can outperform the
naively initialised perfect-model benchmark suggests that, in addition to approximating the underlying system evolution, the D2D model captures part of the state-estimation effect that would otherwise be introduced through data assimilation. Moreover, the fact that the performance of the $128\Delta t$ model converges towards the climatological forecast at longer lead times also indicates that the model represents long-range predictive uncertainty in a physically sensible way, naturally approaching the climatological limit as predictive information is lost. More broadly, these results provide indirect support for the central idea of the D2D framework: for the purpose of skillful probabilistic forecasting, it is not necessary to quantify different sources of uncertainty separately, including initial-condition uncertainty, parameter uncertainty, and model discrepancy. Instead, the D2D model can account for their combined effect within a unified end-to-end probabilistic forecasting framework.

\begin{figure}[htbp]
\centering

\subfloat{%
\includegraphics[width=0.5\linewidth]{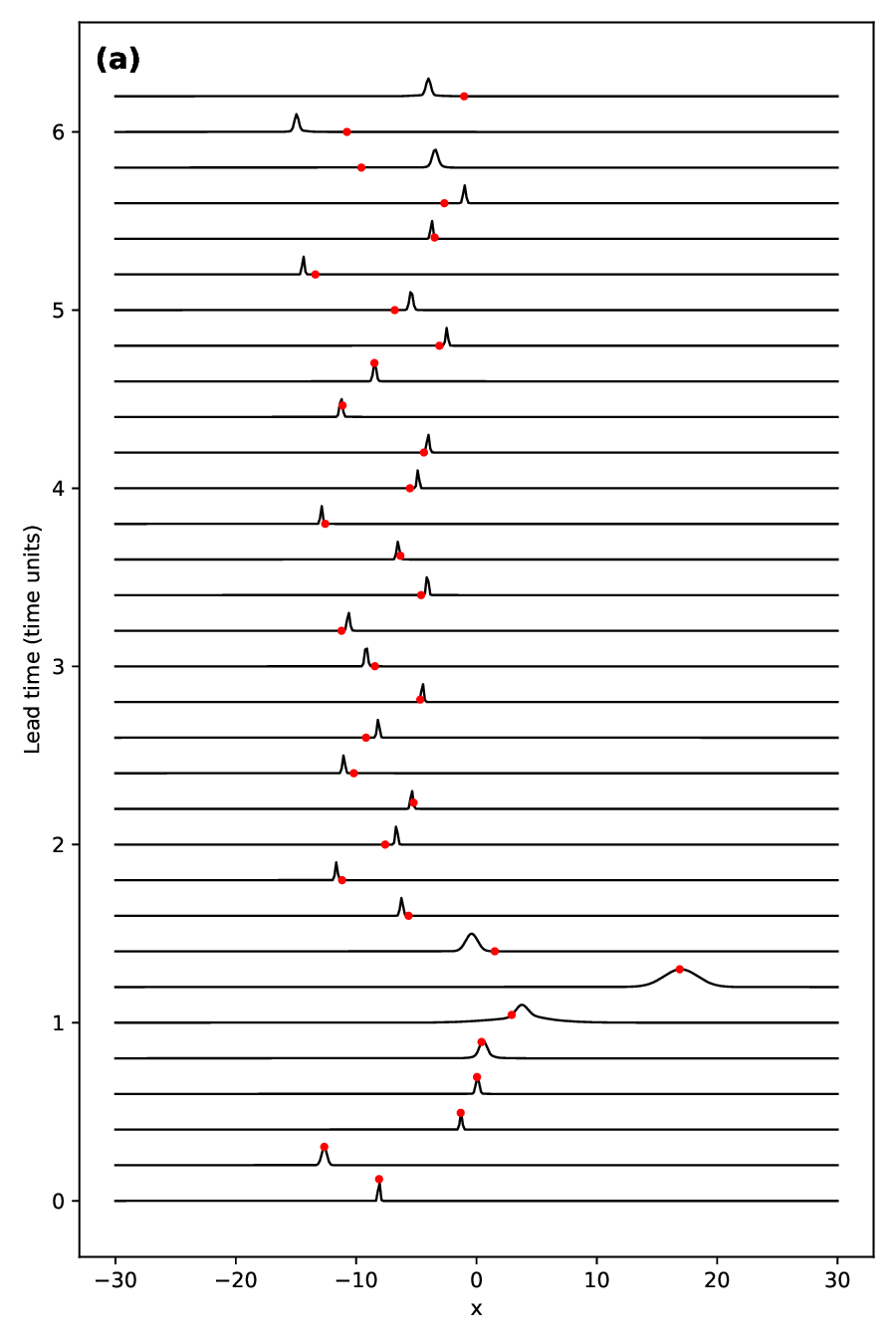}}
\hfill
\subfloat{%
\includegraphics[width=0.5\linewidth]{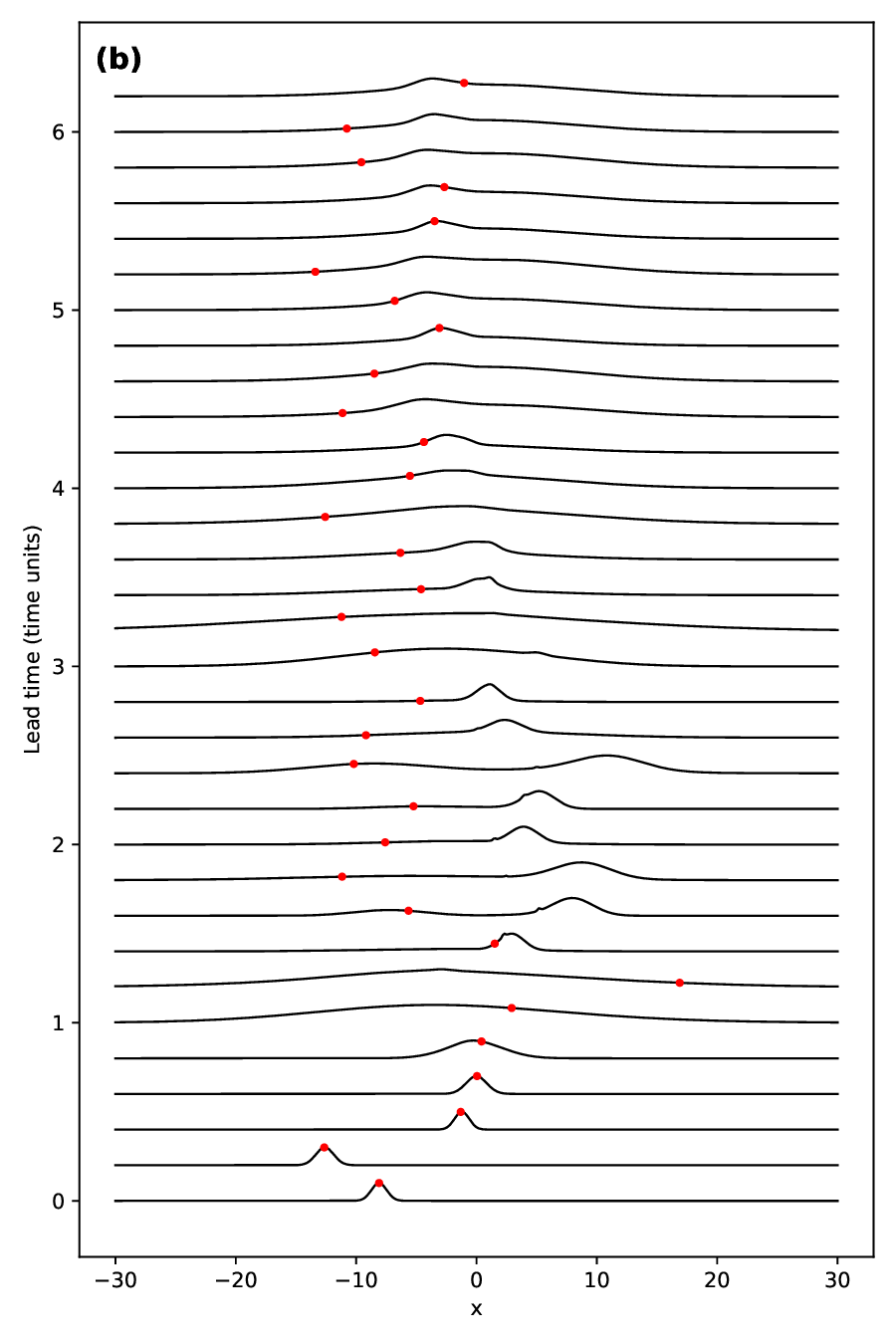}}

\caption{Evolution of the predictive PDF for the \(x\)-variable generated by the \(32\Delta t\) iterative D2D model, using the same initial condition as in Fig.~\ref{fig:evolve}. Panel (a) corresponds to observational noise level \(0.01\), and panel (b) corresponds to observational noise level \(0.08\). }
\label{fig:evolve2}

\end{figure}

An interesting and somewhat counterintuitive observation from Fig.~\ref{fig:loss_it_fig} is that, for models trained with short maximum curriculum lead times, long-range forecast performance improves as the observational noise level increases; for example, this is clearly visible for the $32\Delta t$ model when comparing panels (a) and (d). This can be explained by the fact that, at low noise levels, the short lead time target distributions tend to be relatively narrow and often close to Gaussian, so short range iterative models can achieve good local skill without being sufficiently exposed to non-Gaussian or otherwise complex input-output distribution mappings. As a result, they have limited opportunity during training to learn how to handle the richer distributional structures that become common at longer lead times. By contrast, when the observational noise level is larger, even short lead time target distributions may already exhibit substantial non-Gaussianity, including bimodality. In that setting, short range iterative models are forced to learn more complex probabilistic transformations during training, or at least to produce wider output distributions, which explains their improved robustness at longer lead times. This interpretation is supported by Fig.~\ref{fig:evolve2}, which shows the PDF evolution generated by the \(32\Delta t\) model for the same initial condition under noise levels \(0.01\) and \(0.08\). Under the higher noise setting, the \(32\Delta t\) model generates broader and more non-Gaussian predictive distributions over the entire lead-time range.

 
\begin{figure}[htbp]
\centering

\subfloat{%
\includegraphics[width=0.5\linewidth]{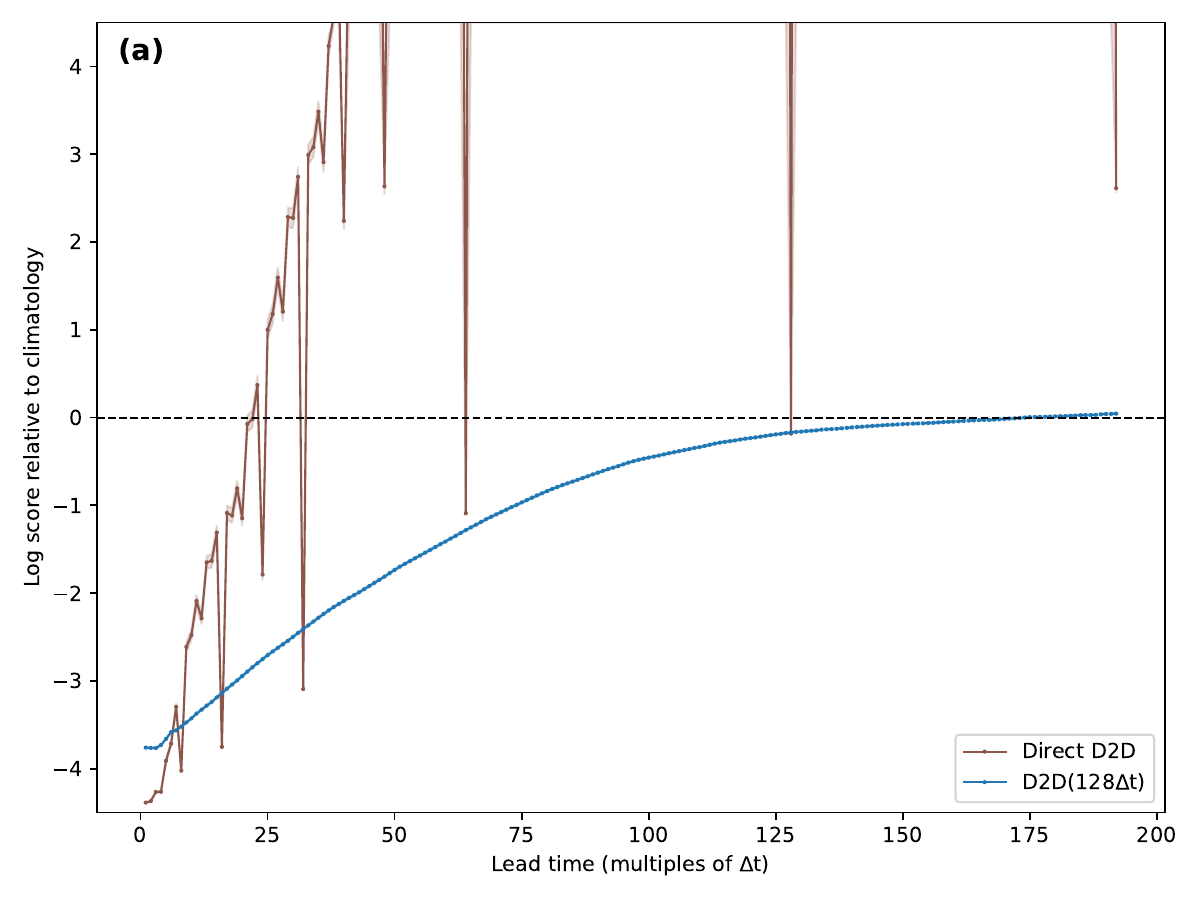}}
\hfill
\subfloat{%
\includegraphics[width=0.5\linewidth]{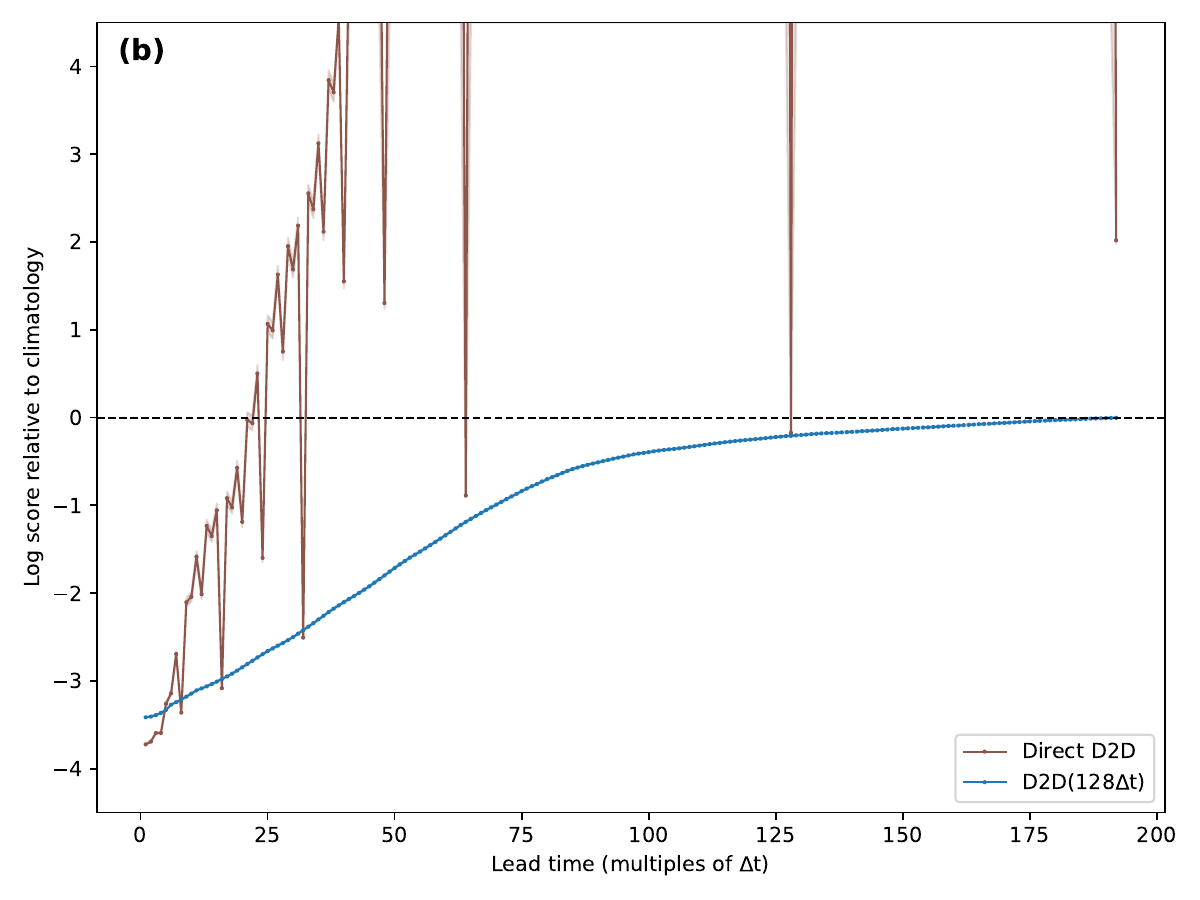}}
\hfill
\subfloat{%
\includegraphics[width=0.5\linewidth]{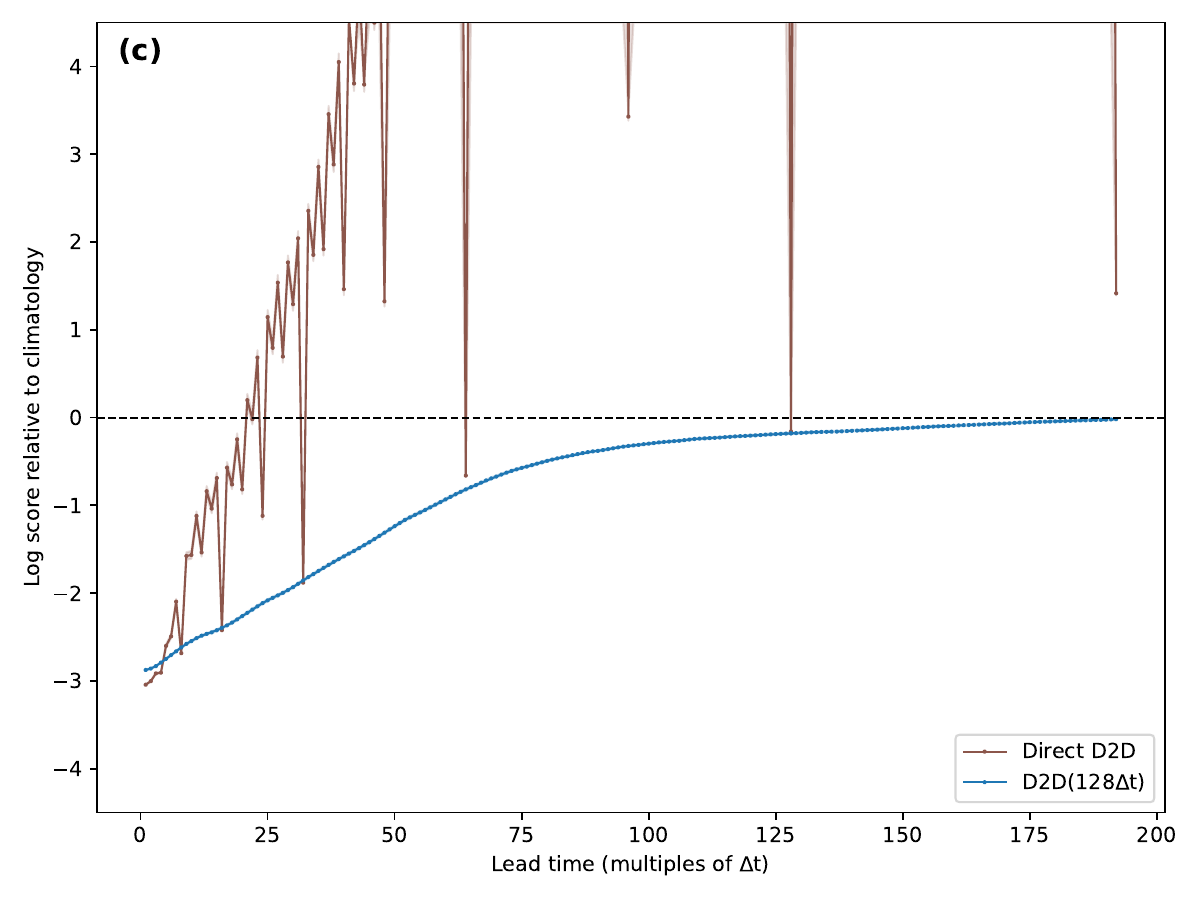}}
\hfill
\subfloat{%
\includegraphics[width=0.5\linewidth]{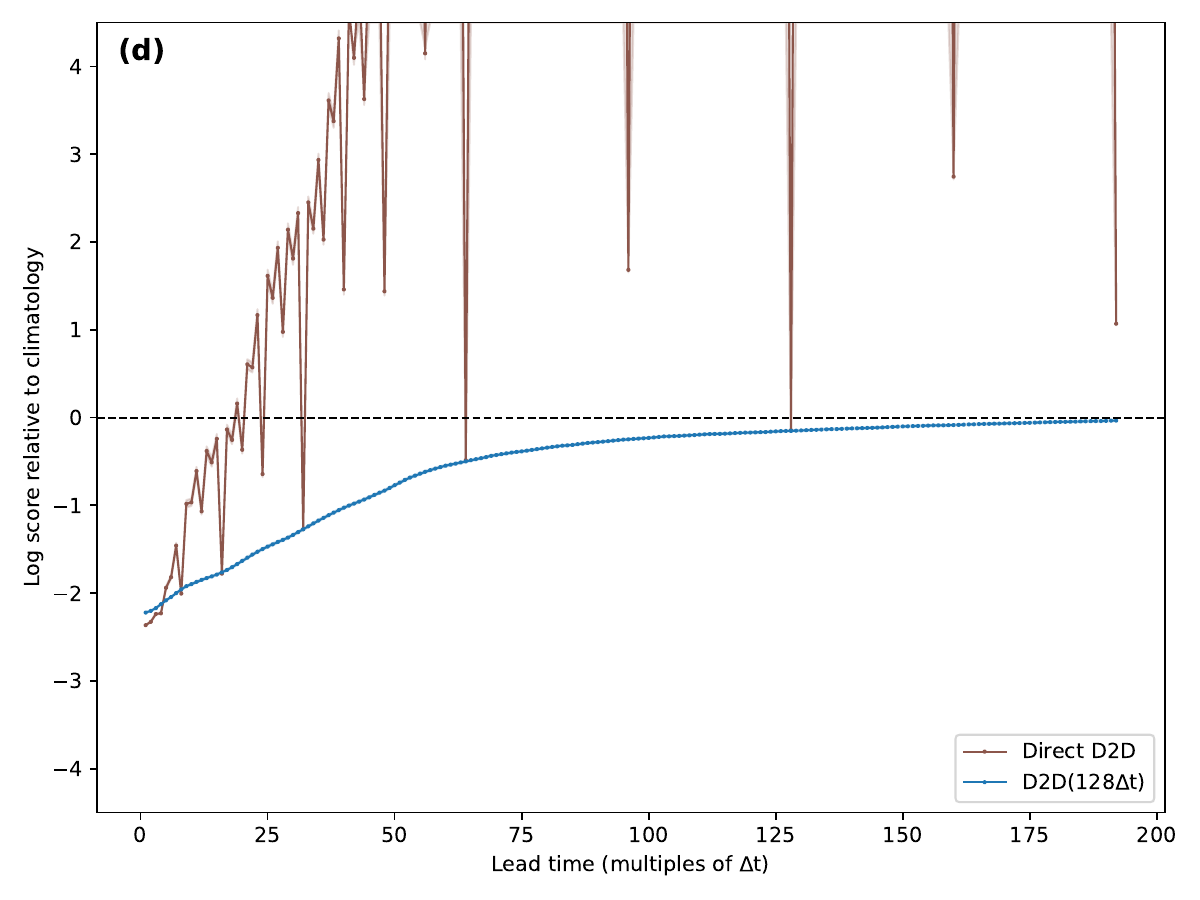}}

\caption{Logarithmic score skill relative to climatology for the direct and iterative D2D strategies under different observational noise levels. Panels (a)--(d) correspond to the same observational noise levels as in Fig.~\ref{fig:loss_it_fig}. The blue curve denotes the iterative D2D model trained with maximum curriculum lead time $128\Delta t$. The red curve denotes the direct-forecast strategy, constructed from multiple direct D2D models trained at fixed lead times $1\Delta t$, $2\Delta t$, $4\Delta t$, $\ldots$, $128\Delta t$ and composed through a hierarchical temporal aggregation strategy to produce forecasts at arbitrary lead times.}
\label{fig:direct_vs_iter}

\end{figure}

To further compare the iterative and direct strategies, Fig.~\ref{fig:direct_vs_iter} shows forecast skill under different observational noise levels, with panels (a)--(d) corresponding to the same four noise settings as in Fig.~\ref{fig:loss_it_fig}. The blue curve denotes the iterative D2D model trained with maximum curriculum lead time $128\Delta t$, while the red curve denotes the direct-forecast strategy. For the direct strategy, a collection of direct D2D models is trained separately for fixed lead times $1\Delta t$, $2\Delta t$, $4\Delta t$, $\ldots$, $128\Delta t$. To mimic the common operational practice in modern neural weather forecasters \citep{bi2023accurate}, multiple fixed lead time direct models are combined through a hierarchical temporal aggregation strategy to produce forecasts at arbitrary lead times. When the target lead time coincides with one of these trained lead times, the corresponding direct model is applied directly. Otherwise, the forecast is produced by sequentially composing the available direct models in a greedy manner, always using the largest admissible lead time first; for example, a forecast at $7\Delta t$ is obtained by applying the $4\Delta t$, $2\Delta t$, and $1\Delta t$ models in succession.

Compared with the iterative D2D model, the direct D2D forecasts constructed through the hierarchical temporal aggregation strategy perform poorly across all four noise levels in Fig.~\ref{fig:direct_vs_iter}, exhibiting large fluctuations and repeated spikes in logarithmic score across lead times. This is because the hierarchical temporal aggregation strategy is not fully consistent with the way the direct models are trained. Each direct model is trained using a Gaussian initial input distribution, whereas the output of a direct forecast, especially at longer lead times, is generally non-Gaussian. When one direct forecast is used as the input to another, the downstream model is therefore applied to distributions that lie outside the regime on which it was trained, naturally leading to unstable and degraded forecast performance.

For the individually trained direct models at fixed lead times $1\Delta t$, $2\Delta t$, $4\Delta t$, $\ldots$, $128\Delta t$, an advantage over the iterative $128\Delta t$ model is observed only at very short lead times. Even this short-range advantage diminishes as the observational noise level increases, suggesting that the iterative model benefits more from information that would otherwise need to be incorporated through data assimilation, because its longer recursive training allows it to exploit a richer range of dynamical information.




\section{Summary and Discussion}

This work proposes a distribution-to-distribution (D2D) probabilistic forecasting framework for dynamical systems, in which predictive distributions are treated as the primary objects of forecasting. A key innovation of the proposed D2D framework is its distributional encoding and decoding structure, which allows existing neural forecasting modules to operate directly on predictive distributions, thereby enabling recursive uncertainty propagation and direct optimisation under the logarithmic score. 
More specifically, the proposed framework provides a practical route for learning the dynamical evolution of predictive distributions using neural networks, thereby offering a computationally tractable alternative to the direct numerical evolution of predictive densities, which can be prohibitive for nonlinear dynamical systems; it makes probabilistic forecasting an intrinsic part of the model rather than something constructed externally through ensembles or post hoc calibration; and it offers a unified approach to uncertainty quantification, in which different sources of uncertainty need not always be represented separately; instead, the framework can learn their combined effective impact within a single end-to-end probabilistic model, with probabilistic forecast skill providing the defining criterion for the quality of uncertainty quantification.

Experiments on the Lorenz63 system provide a proof-of-concept demonstration that the proposed framework can learn nontrivial distributional evolution under nonlinear chaotic dynamics. The iterative D2D model captures the deformation of predictive distributions over time and yields skillful probabilistic forecasts without explicit ensemble simulation. Quantitative evaluation using logarithmic score shows that models trained with longer recursive curriculum lead times generally achieve more robust performance across a broad range of forecast lead times, whereas models trained only over short lead times deteriorate rapidly once applied beyond the regime encountered during training. The iterative strategy also substantially outperforms the direct strategy when forecasts are extended or composed across multiple lead times, underscoring the importance of learning a dynamically consistent distributional evolution operator rather than a collection of lead-time-specific predictors. Furthermore, iterative D2D models are shown to be capable of competing with, and in some cases even outperforming, a naively initialised perfect model benchmark, suggesting that the framework captures not only the underlying system evolution but also part of the predictive effect of dynamically consistent initialisation normally associated with data assimilation.

Although the experiments in this paper use marginal predictive distributions, this is not a structural limitation of the D2D framework. A joint version can be obtained by combining multivariate MDN outputs with multivariate KME inputs. The principal challenge is scalability rather than formulation: full covariance mixtures introduce quadratic parameter growth with the state dimension, and joint embedding centres must be chosen carefully in high-dimensional systems.

In the present marginal implementation, however, the model is not restricted to isolated marginal distributions at a single time. Instead, the forecasting module receives a temporal window of embedded marginal distributions across all state variables. This allows the model to exploit information arising from
both temporal evolution and cross-variable dynamical coupling, even though the contemporaneous joint predictive distribution is not represented explicitly. This is consistent with the intuition of Takens-type delay embedding \citep{takens2006detecting}, according to which temporal histories of partial observations can contain substantial information about the underlying dynamical state. Although this does not imply recovery of the full
joint predictive distribution, it allows the marginal D2D model to exploit aspects of the dependence structure that are relevant to the evolution of the individual predictive marginals. In this sense, the use of temporal information is also related to the role of data assimilation, where observations collected
over time are combined with dynamical information to improve inference about the underlying system state. The marginal formulation therefore provides a computationally efficient compromise, while explicit joint distributional modelling remains an important extension for representing cross-variable
predictive dependence in full.


More broadly, one of the deeper reasons for the effectiveness of the D2D framework is that it represents uncertainty directly in the form of probabilistic forecasts, thereby enabling both training and evaluation to be carried out using the logarithmic score, which provides the unique principled standard for assessing probabilistic forecast skill. This may also help explain, at least in part, why neural networks have achieved greater success in classification problems, where cross-entropy provides a natural probabilistic objective, than in continuous-valued forecasting and regression, where no equally universal standard is as consistently adopted. From this perspective, the D2D framework points toward a broader direction for AI-based forecasting, in which the direct probabilistic representation and quantification of uncertainty, together with its evaluation through logarithmic score, become central design principles.


\section*{CRediT authorship contribution statement}

\textbf{Tianlin Yang:} Data curation, Formal analysis, Investigation, Methodology, Resources, Software, Validation, Visualization, Writing - original draft, Writing - review \& editing.

\textbf{Hailiang Du:} Conceptualization, Data curation, Formal analysis, Funding acquisition, Investigation, Methodology, Project administration, Resources, Supervision, Validation, Visualization, Writing - original draft, Writing - review \& editing.

\textbf{Louis Aslett:} Supervision, Writing - review \& editing.

\section*{Declaration of competing interest}
The authors declare that they have no known competing financial interests or personal relationships that could have appeared to influence the work reported in this paper.

\section*{Data availability}
Data and code will be made available on reasonable request.

\section*{Acknowledgments}
Hailiang Du acknowledges support from the EPSRC-funded project Virtual Power Plant with Artificial Intelligence for Resilience and Decarbonisation (Grant No. EP/Y005376/1) for the completion of this work. Louis Aslett was supported by the EPSRC research grant "Pooling INference and COmbining Distributions Exactly: A Bayesian approach (PINCODE)" (EP/X028100/1), and by the UKRI grant, "On intelligenCE And Networks (OCEAN)" (EP/Y014650/1).







 \bibliographystyle{elsarticle-num} 
 \bibliography{bibliography}






\end{document}